\documentclass[a4paper,11pt]{llncs}
\usepackage{graphicx}

\usepackage{tikz}
\usepackage{comment}
\usepackage{amsmath,amssymb} 
\usepackage{color}

\usepackage[accsupp]{axessibility}  

\usepackage{subcaption} 
\usepackage{amsmath   } 
\usepackage{bm        } 
\usepackage{url       } 
\usepackage{comment   } 
\usepackage{algorithm } 
\usepackage{algorithmic}
\usepackage{booktabs  } 
\def\vs{\emph{vs. }}    
\def\eg{\emph{e.g. }}   

\begin{document}

\title{G2L: A Geometric Approach for Generating Pseudo-labels that Improve Transfer Learning}


\author{John R. Kender\inst{1} \and  Bishwaranjan Bhattacharjee\inst{2}
\and Parijat Dube\inst{2} \and Brian M. Belgodere\inst{2}}

\institute{Columbia University, New York, NY 10027,\\
\email{jrk@cs.columbia.edu},
\and
IBM Research, Yorktown Heights, NY 10598\\
\email{\{bhatta, pdube, bmbelgod\}@us.ibm.com}
}

\maketitle

\begin{abstract}
Transfer learning is a deep-learning technique that ameliorates the problem of learning when human-annotated labels are expensive and limited.  In place of such labels, it uses instead the previously trained weights from a well-chosen source model as the initial weights for the training of a base model for a new target dataset.  We demonstrate a novel but general technique for automatically creating such source models.  We generate pseudo-labels according to an efficient and extensible algorithm that is based on a classical result from the geometry of high dimensions, the Cayley-Menger determinant.  This G2L (``geometry to label'') method incrementally builds up pseudo-labels using a greedy computation of hypervolume content.  We demonstrate that the method is tunable with respect to expected accuracy, which can be forecast by an information-theoretic measure of dataset similarity (divergence) between source and target.  The results of 280 experiments show that this mechanical technique generates base models that have similar or better transferability compared to a baseline of models trained on extensively human-annotated ImageNet1K labels, yielding an overall error decrease of 0.43\%, and an error decrease in 4 out of 5 divergent datasets tested.
\end{abstract}


\section{Introduction}
\label{introduction}

In the field of supervised deep learning, one often ends up with very little labeled training data. To alleviate this problem, a well-known technique used is Transfer Learning~\cite{NIPS2014_5347}. It uses trained weights from a source model as the initial weights for training a target dataset. A well-chosen source with a large amount of labeled data leads to significant improvement in accuracy for the target.

The task we address is the scenario of providing an efficient machine learning service for clients whose data, particularly image data, varies greatly from the ``natural scenes'' that typically populate the image databases on which off-the-shelf classifiers are trained.  
Typically, these clients will present a rather small dataset taken from specialized environments, often industrial, which have been captured under unforeseen circumstances and parameters. 
Our method creates pseudo-labels for this data by determining their geometric relationships to the feature space of existing labeled data.
 
In this work, we therefore develop a content-aware labeling technique.
First, we take data points such as images, and compute labels for them by calculating distances of these data points from a set of named anchor data points representing known and labeled categories, 
like $animal$, $plant$, $tool$, etc.  Second, pseudo-labels are then constructed for the incoming data points based on these distances, or, more accurately, based on ``contents'', which is the high-dimensional generalization of distances,
calculated using a geometric approach.  Each pseudo-label then consists of a sequence of semantically descriptive names: for example, $\langle tool, plant \rangle$, could be the pseudo-label for data from a previously unseen category like $rake$.
Finally, we train a source model using these automatically labeled labels.  
We show that when an incoming dataset has a high divergence from the data on which a classifier has been trained, this method helps to focus transfer learning on the most relevant aspects of the training data, and the resultant model is more accurate than the original ``vanilla'' classifier.

Applying this G2L (``geometry to label'') method, we evaluate workloads from the Visual Decathlon ~\cite{odeca} and other labeled datasets, comparing how well our pseudo-labeling scheme performs in generating sources for transfer learning against ImageNet1K data ~\cite{ILSVRC15} using standard human annotated labels.
We show that our purely mechanical approach wins in many circumstances and that we can specify those circumstances.

Our contributions are: 
(1) a disciplined extensible algorithm to create semantically interpretable pseudo-labels, derived from methods in the geometry of high dimensions;
(2) an analysis of the effects of algorithm hyperparameters on the amount and quality of these pseudo-labels;
(3) a characterization of the relationship of source-to-target similarity (divergence) \vs pseudo-label variety, optimal transfer learning rate, and final accuracy; and
(4) a discussion and visualizations of 280 experiments on a wide variety of transfer tasks.

\section{Related Work}
\label{related:sec}
Several well-established approaches attempt to assign labels to unlabeled images automatically. Some utilize feature clusters to predict labels~\cite{Jeon:2003:AIA:860435.860459} or augment image data with linguistic constraints from sources such as WordNet~\cite{barnard2003matching,jin2004effective}. They address tasks by pretraining models using larger unlabeled data-sets. 

Pretraining approaches have also improved results when attempting a target task with a limited amount of accurately labeled training data~\cite{Mahajan}, by using \emph{weakly} labeled data, such as social media hashtags, which are very plentiful. Again, however, effectiveness only appears to grow as the log of the image count.

Other approaches use generative models such as GANs~\cite{DBLP:journals/corr/RadfordMC15} to explore and refine category boundaries between data clusters, which exploit the rich statistical structures of both real and generated examples, sometimes augmented with labels or linguistic constraints.  These automatic approaches use the structures present in large unlabeled datasets to extend the expressivity of known labels and augment the raw size of training sets. 

More broadly, various approaches attempt to learn a representation of a class of data and later use that representation in service of a target task. 
For example, \cite{Hsu2018} clustered images in an embedding space and developed a meta-learner to find classifications which distinguished various clusters within this embedding. 
Other approaches to mapping the feature space have used autoencoders~\cite{DBLP:journals/corr/abs-1811-00473}.  
Knowledge distillation techniques~\cite{mirzadeh2020improved} teach a student an accurate but more compact feature space, similar in spirit to the work presented here.



The above existing literature attempts to find hybrid approaches that find productive ways to leverage machine-learned distributions of examples to find new ways of characterizing unlabeled data. The current work presents a novel approach in this domain.

\section{General Approach}
\label{sec:approach}
In this section we give an overview of the conceptual flow of our method, discuss the feature space of one application of it to a computer vision task, and suggest a real-world analogy that illustrates some of the semantic considerations behind its central geometry-based algorithm.


\subsection{Labeling Method}

Generating rich pseudo-labels from models trained on distributionally similar data involves a trade off between an expressive, long label, and a generalizable, short label. Longer labels carry more information about similarity between previous models and the target image, and differences between the previously trained models could be critical for adequately labeling new examples. For example, an incoming set of data including pictures of household objects might be well described by combining the labels of “tool, fabric, furniture.” 

However, domains that possess substantial differences from previous data might be better defined by the magnitude and direction of such a difference. For example, a ``flower'' dataset would share some features with ``plant,'' but it is perhaps better defined by statements such as ``flowers are very unlike furniture''.  In other, ambiguous cases, negative features may be necessary to distinguish between overlapping cases: a suit of armor might have similarities with the body shapes of people but could be contrasted with these categories by its dissimilarity with ``sport,'' a category otherwise close to ``person.''

\subsection{Labeling Example}

We illustrate our method using a specific case study involving images, and with source datasets created by vertically partitioning ImageNet22K \cite{imagenet22k} along these distinct subtrees: $animal$, $building$, $fabric$, $food$, $fruit$, $fungus$, $furniture$, $garment$, $music$, $nature$, $person$, $plant$, $sport$, $tool$, $tree$, $weapon$, illustrated later in Fig.~\ref{fig:confusion16}.
These subtrees vary in their number of images (from 103K images for $weapon$ to 2,783K images for $animal$) and in their number of classes (from 138 for $weapon$ to 4,040K for $plant$).
These 16 subtrees were used since they were easy to partition from Imagenet22K, but our method could also be used with any other labeled data ontology.

We represent each such source dataset by a single average feature vector. 
This study generates this vector
from the second to last layer of a reference VGG16 \cite{vgg16} model trained on ImageNet1K,
with the average taken over 25\% of all the images in the dataset.

To label a new image, we first calculate its own feature vector, then compute its Euclidean distance from each of the representatives of the datasets. 
Together with other geometric computations in this high dimensional space, these distance measures are then used for a full pseudo-labeling process described in Sec.~\ref{sec:geometry}.

\subsection{An Analogy}
Our G2L approach for pseudo-labeling an image can be understood as being similar to the ``Blind Men and the Elephant'' parable, where blind men, who have never learned about an elephant, try to categorize an elephant just by touching it, then relating it to something that they already know. 
Their categorizations of Elephant then include Fan (ear), Rope (tail), Snake (trunk), Spear (tusk), Tree (leg), and Wall (flank).  Basically, by touching and feeling an elephant, the blind men are measuring its closeness to things known by them.  Our approach also measures the closeness of an unknown image, in feature space, to existing known categories and then generates a pseudo-label for it.

\subsection{Analogy Extended}

However, additionally, our work extends this analogy in three critical ways.  First, we also compare unknown images to the existing categories that are {\em farthest} from them: the elephant is ``not Feather''.  Second, we observe a strong predictive relationship between (a) the measurement of the similarity of unknown imagery to existing categories, and (b) the computation of the {\em number} of pseudo-labels necessary to derive good transfer performance: the elephant could also use  ``Curtain, Leather, Bark, Mud''.  Third, we also observe a strong predictive relationship between measured similarity and optimal {\em learning rate}: the elephant is most easily described starting from ``Manatee''.  The significance and computational advantages of these extensions are detailed in Sec.~\ref{sec:geometry} and \ref{sec:observations}.

\section{Geometric Pseudo-labels} 
\label{sec:geometry}

\newcommand{\T}[1]{\mathtt{#1}} 


In this section we present the main algorithm for generating semantically meaningful geometric pseudo-labels.  We first start with five motivating principles and some necessary math background.  Then we walk through the algorithm, and discuss some of its properties and results.  

\subsection{Motivation}
\label{sub:motivation}

Pseudo-labels for a target dataset can be generated by using a
large {\em labeled dataset} organized within a semantic
hierarchy such as ImageNet22K, and an off-the-shelf {\em robust classifier},
such as VGG16 trained on ImageNet1K. (The robust classifier need
not be trained on the same labeled dataset.) 
Our algorithm exploits these two tools in a way that promotes
five desirable properties for pseudo-labels, to ensure that the pseudo-labels have an easily understandable meaning, and that their computation are reasonable efficient.

First, the pseudo-labels should be easily interpretable to humans.
As an example, we can start by partitioning ImageNet22K into 16 non-intersecting sets, each of which carries the name of an object category, such as in Sec.~\ref{sec:approach}.

Second, the pseudo-labels should therefore also follow a simple grammar.
We refer to the 16 subsets that comprise the above partition as the source
subsets.  Then, a pseudo-label for an incoming target data item can be
defined as the concatenation of some number of source subset names,
such as the sequence
$\langle person, music, tool \rangle$.  This 
produces an informative natural language description of the incoming target data item.

Third, to make comparisons possible, these pseudo-labels should be geometrically interpretable within the space of feature vectors.
Distances between pseudo-labels can be defined by various
metrics: $L^1$
(city-block), $L^2$ (Euclidean), $\sqrt{JS}$ (the square root of
Jenson-Shannon divergence~\cite{1207388}), or others. 
This can be tricky; the 
feature vector spaces used in machine learning are difficult to
visualize, and such high-dimensional spaces generate geometric
paradoxes even at relatively low dimensions~\cite{hayes2012adventure}.
For example, each feature vector of a dataset is very likely to be on
the convex hull of that dataset's representation in that
space~\cite{yousefzadeh2021deep}.  
Nevertheless, by methods like that of barycentric
coordinates~\cite{khan2015linear}, 
particular ``anchor''
vectors can be used to represent regions within these spaces.

\begin{figure*}[!ht] 
\begin{subfigure}{.24\linewidth}
  \centering
  \includegraphics[width=\linewidth]{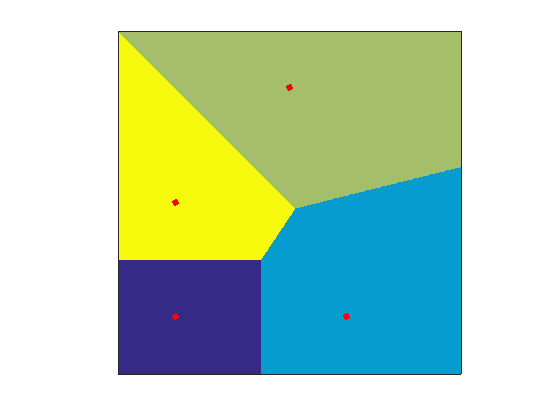}
  \caption{$1^{st}$-order}
\end{subfigure}
\begin{subfigure}{.24\linewidth}
  \centering
  \includegraphics[width=\linewidth]{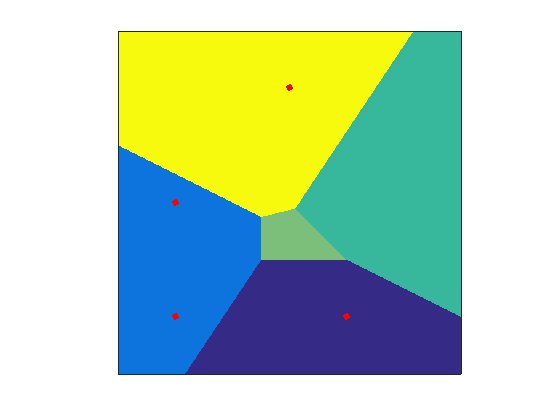}
  \caption{$2^{nd}$-order}
\end{subfigure}
\begin{subfigure}{.24\linewidth}
  \centering
  \includegraphics[width=\linewidth]{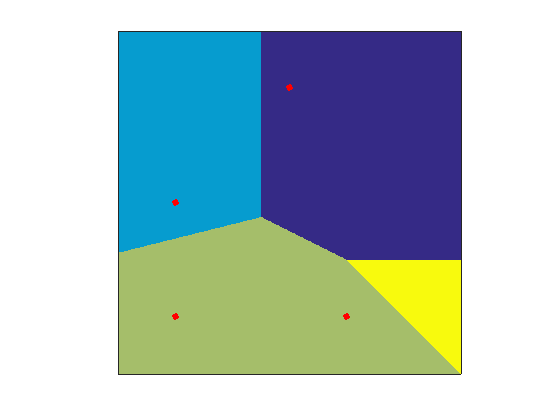}
  \caption{$3^{rd}$-order}
\end{subfigure}
\begin{subfigure}{.24\linewidth}
  \centering
  \includegraphics[width=\linewidth]{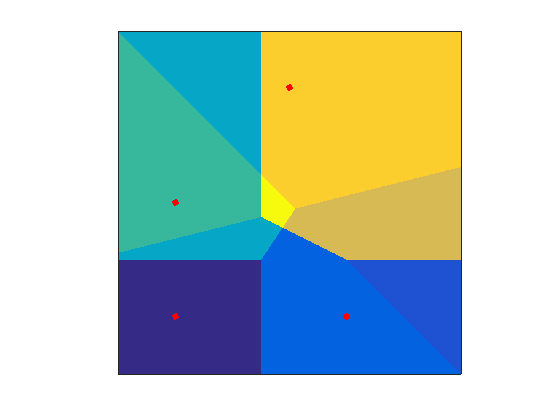}
  \caption{$1^{st}$ $\cap$ $3^{rd}$-order}
\end{subfigure}
\caption{Voronoi tessellation regions in two-dimensions generated by the $N{=}4$ points shown in red.  Colored regions depict equivalence classes of points that share: (a) ``closest point'' (policy $\T{c}$), (b) ``top 2 closest points'', (c) ``farthest point'' (policy $\T{f}$), (d) ``closest and farthest points'' (policy $\T{C}$ or $\T{F}$), the intersection of (a) and (c).  Policies are explained in Sec.~\ref{subsec:pseudocreate}}
\label{fig:voronoi}
\end{figure*}

Fourth, the computation of pseudo-labels should leverage known efficient geometric algorithms that localize incoming data.
Metrics defined over these spaces can be used to partition the space
into cells that form equivalence classes of locations based on
individual anchor points (``$1^{st}$-order Voronoi diagram'').  These
locations are characterizable by geometric properties such as ``the
nearest point to this cell is $P$''; see Fig.~\ref{fig:voronoi}(a).  These
cells can be efficiently determined~\cite{dwyer1991higher}.
Further,
these metrics can also partition the space into cells
that form equivalence classes of locations based on {\em
sets} of points (``$n^{th}$-order Voronoi diagram''), characterizable
by geometric properties such as ``the $n$-nearest points to this cell
are $\{P_1, P_2, \dots, P_n\}$; see Fig.~\ref{fig:voronoi}(b).  The
extreme case for $N$ points is the $(N{-}1)^{th}$-order partition
(``farthest-point Voronoi diagram''); see Fig.~\ref{fig:voronoi}(c).
However, general farthest-point algorithms are provably hard, and the
only efficient algorithms are approximate~\cite{pagh2015approximate}.

Fifth, the concepts of lengths and distances in these spaces
should be further generalized to all measures of higher-dimensional
``content'', following the progression of
polytopes~\cite{brondsted2012introduction}, as point, length, area,
volume, hypervolume, etc., to arbitrarily high dimension.  Elegant algorithms exist for computing
such content, in particular, the Cayley-Menger
determinant~\cite{sommerville1958}.

Therefore, pulling these observations together, we seek to devise a method that composes
a small number of semantically-named anchor vectors derived from the
source datasets, into a sequence that defines location
descriptions for target data items, based on the generalization of
closest and farthest (Voronoi) distances into minimal and maximal
(Cayley-Menger) contents.  These location descriptions then become the
pseudo-labels for the incoming data.  

\subsection{Mathematical Foundation}
Some necessary mathematical preliminaries now follow.

\textbf{Cayley-Menger determinant.}
Our method depends on the generalization of the concept of a single distance between a target
and a single source, to that of the content of a $d$-dimensional
simplex defined by the target and $d$ sources.

The computation of content is a well-studied algorithm based on the
Cayley-Menger determinant (``$CM$'').
The
determinant itself generalizes several earlier classic algorithms,
including the familiar Heron formula for the area of a triangle, and the less
familiar Piero formula for computing the volume of a
tetrahedron. 

\textbf{Cayley-Menger computation.}
For an $d$-simplex, composed of $d{+}1$ anchors, the math to compute
content $C_d$ proceeds in three steps.  The derivation of these steps
is tedious, and is explained in~\cite{blumenthal1970theory}.  

First, it forms $\bm{M}_d$, a particular symmetric
$(d{+}2){\times}(d{+}2)$ matrix.  It incorporates a symmetric
submatrix that expresses the {\em squares} of all pair-wise distances,
that is, $D_{i,j}{=}distance(i,j)^2$.  

\begin{equation*}\setlength\arraycolsep{2pt}
\bm{M}_d=
\begin{bmatrix}
 0         & 1         & 1         & 1         & 1         & \cdots    & 1       \\
 1         & 0         & D_{0,1}   & D_{0,2}   & D_{0,3}   & \cdots    & D_{0,n} \\
 1         & D_{1,0}   & 0         & D_{1,2}   & D_{1,3}   & \cdots    & D_{1,n} \\
 1         & D_{2,0}   & D_{2,1}   & 0         & D_{2,3}   & \cdots    & D_{2,n} \\
 1         & D_{3,0}   & D_{3,1}   & D_{3,2}   & 0         & \cdots    & D_{3,n} \\
 \vdots    & \vdots    & \vdots    & \vdots    & \vdots    & \ddots    & \vdots  \\
 1         & D_{n,0}   & D_{n,1}   & D_{n,2}   & D_{n,3}   & \cdots    & 0       \\
\end{bmatrix}
\label{eq:M}
\end{equation*}

Second, it computes the coefficient $a_d=(-1)^{d+1} 2^d (d!)^2$, which records the effects
that various matrix operations have had on the determinant of
$\bm{M}_d$, during its simplification from more complex geometric
volume computations into its above form.  This coefficient also
defines the integer sequence A055546 at \cite{sloane2003line}, where
it has an imaginative interpretation involving roller coasters.

Third, it solves for the value of $C_d$ implicitly expressed by $a_d C^2_d=\det{\bm{M}_d}$.

The complexity of computing the determinant, by the usual and
reasonably efficient method of LU decomposition, is
$\mathcal{O}(d^3)$.  No simpler approach involving the reuse of
previously computed subdeterminants appears feasible, as the
determinant has been proven to be irreducible for dimensions greater
than $3$~\cite{d2005cayley}.  
Nevertheless, in the context of our overall machine learning problem, this cost has proven to be negligible with respect to training costs.

\subsection{Pseudo-label Creation}
\label{subsec:pseudocreate}

Now, we give an overview of the algorithm.  An incoming data point is compared at each step against a collection of named category anchor points.  The name of the anchor point that minimizes (or maximizes, depending on a policy) its distance to the incoming point is chosen as the first component of an evolving sequence of names.  Thereafter, the process repeats, and at each step the sequence is extended with the name of the anchor point that best extremizes the content---the area, volume, hypervolume, etc.--- of the evolving polytope formed by these selected points.  After a stopping criteria, this sequence gives the pseudo-label.

The full G2L algorithm is summarized in Algorithm~\ref{alg:pseudo}.  The algorithm
requires a number of hyperparameters that are set by experiment.  An
example is shown for each of these choices, in the pseudocode of the
precondition (``Require'') preamble.  These examples use image
classification as the domain, and record the exact configuration
that is used in the experiments reported in Fig.~\ref{fig:winloss1}.

Most of the parameters (and hyperparameters) are straightforward.  The
indicator $Layer$ is the choice of a particular layer within the data
representation of $f$, usually but not necessarily the second-last.
The function $Met$ is the choice of a distance function that has been
derived from an inner product, as required for the derivation of
$CM$. The remaining three hyperparameters are more complex.

\begin{figure}[ht]
\centering
\includegraphics[scale=0.75]{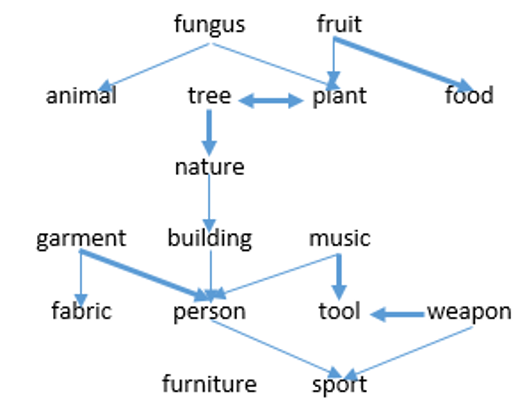}
\caption{The 16 sources, using 1NN under Euclidean metric to define ``closest''.  Arrow $a{\rightarrow}b$ means ``$a$ includes a $b$ subcluster''; thickness is subcluster weight.}
\label{fig:confusion16}
\end{figure}

\textbf{Parameters needing explanation.}
The method $Aggr$ is the choice of an aggregation method that
represents a set of $Layer$ vectors in a sparser form.  This can be as
trivial as using a single mean vector, or as more elaborate as using a
set of representatives derived from clustering methods.  For example,
as Fig.~\ref{fig:confusion16} suggests, the source $food$ is probably
adequately represented by a single aggregate vector, but the source
$fruit$ probably is better represented by a pair of aggregate vectors
$fruit_{plant}$ and $fruit_{food}$.

The integer $d_{max}$ determines the number of dimensions to be
explored using $CM$ during the creation of the output pseudo-label name
sequences.  It also bounds the length of the pseudo-label name sequence $pls_i$, by
$d_{max} \le \lvert pls_i \rvert \le 2d_{max}$.  The exact length of
$pls_i$, which is constant over a given execution of the complete
algorithm, is determined by $Pol$.

\textbf{Extremizing policies.}
The extrema decision sequence $Pol$, and its summarizing notation, are
best explained by a walkthrough of the algorithm.

At $d{=}1$, the algorithm considers the length of the line (the
$1$-simplex) formed from the target data item $t_i$, and a
representative vector $sour_{j,k}$ from the source representation
$Sour_j$. (If $Aggr$ was a simple mean, then each $Sour_j$ will be a
singleton set.)  Each $sour_{j,k}$ is examined, and the content (here,
the length), computed by $CM$, is recorded in $cont_{i,j,k}$.

Now we can choose the first dimension's extremizing pseudo-label
sequence $pls_1$ for $t_i$, from one of four short sequences:
(1)
the source name of the closest vector, if $Pol$ starts with $\langle
c \rangle$, as shown in Fig.~\ref{fig:voronoi}(a); or (2) the source name
of farthest vector, if $Pol$ starts with $\langle f \rangle$, as shown
in Fig.~\ref{fig:voronoi}(c); or (3) the source name of the closest vector
followed by the source name of the farthest vector, if $Pol$ starts
with $\langle C \rangle$, as shown in Fig.~\ref{fig:voronoi}(d); or (4)
the source name of the farthest vector followed by the source name of
the closest vector, if $Pol$ starts with $\langle F \rangle$, as shown
in Fig.~\ref{fig:voronoi}(d) again (the repeat is expected in this case).

For example, if $Pol{=}\langle c \rangle$, one possible pseudo-label
$pls_1$ for a particular $t_i$ could be the sequence $\langle
fruit_{food} \rangle$.  Whereas, if $Pol{=}\langle F \rangle$, it
could be $\langle fungus$, $fruit_{food} \rangle$ instead.  The four
choices of extremizing policy at any dimension are therefore captured
by the quaternary alphabet $\{c, f, C, F\}$.  And in particular, the
policy $\langle C \rangle$ is the special case already explored in
prior work~\cite{Dube_2019}, which forms pseudo-labels
consisting of the names of $\langle closest, farthest \rangle$ pairs.

Proceeding to $d{=}2$, the algorithm considers the areas, computed by
$CM$, of the triangle ($2$-simplex) formed by the target data item
$t_i$, a representative vector $sour_{j,k}$, and a single prior
extremizing vector, chosen according to the first dimension's
policy. This single vector would be the length-minimizing vector if
the policy had been $\langle c \rangle$ or $\langle C \rangle$; or the
length-maximizing vector if the policy had been $\langle f \rangle$ or
$\langle F \rangle$.  At this point, again we can efficiently choose
one of four short sequences that capture the names of the
area-extremizing sources for this dimension's pseudo-label, which we
then append to the evolving sequence $pls_i$.

At two dimensions, there are therefore 16 total policies, ranging from
$\langle c, c \rangle$ to $\langle F, F \rangle$.  These 16 policies
create 4 different name sequences of length 2, 8 different name
sequences of length 3, and 4 different name sequences of length 4.  By
establishing and solving straightforward recurrence relations that are
similar to those describing Pascal's triangle, we find that the number
of possible name sequences at dimension $d$ with length $l$ is given
by $P(d,l){=}2^l \cdot \binom{d}{l{-}d}$, and that the total possible
sequences at dimension $d$ is given by $\sum_{l} P(d,l){=}4^d$.

The algorithm proceeds likewise for each higher dimension, up to
$d_{max}$, by first building simplices that extend the prior
dimension's simplex, and then selecting names according to this higher
dimension's policy.

\begin{algorithm}[hbt!]
\caption{G2L pseudo-label algorithm, in pseudocode}
\label{alg:pseudo}

\begin{algorithmic}[1]
\REQUIRE $Tar \gets$ target dataset of data items, $Tar{=}\cup t_i$
\REQUIRE $Full \gets$ semantically-partionable labeled dataset \COMMENT{\eg ImageNet22K}
\REQUIRE $Part \gets$ partition of $Full$, $Part{=}\cup \{P_j\}$
\REQUIRE $f \gets$ classifier \COMMENT{\eg VGG16 on Imagenet1K}
\REQUIRE $Layer \gets$ feature vector layer \COMMENT{\eg second-last}
\REQUIRE $Aggr \gets$ feature vector aggregator \COMMENT{\eg mean}
\REQUIRE $Met \gets$ feature vector metric \COMMENT{\eg Euclidean}
\REQUIRE $d_{max} \gets$ max simplex dimension \COMMENT{\eg $4$}
\REQUIRE $Pol \gets$ extrema decision sequence \COMMENT{\eg $\T{Cfff}$}
\ENSURE pseudo-label sequence $pls_i$ for each $t_i$
\STATE {INITIALIZATION}
\FOR {each data item $t_i \in Tar$}
   \STATE {represent $t_i$ by $vert_i \gets$ $Layer$ vector of $t_i$ within $f$}
\ENDFOR
\FOR {each subset $P_j \in Part$}
   \STATE {represent $P_j$ by set $Sour_j{=}\cup \{sour_{j,k}\} \gets$ aggregation of $Layer$ vectors of $P_j$ within $f$, using $Aggr$}
\ENDFOR
\STATE {PROCESS}
\FOR {each $t_i \in Tar$ }
   \STATE {$pls_i \gets \langle \rangle$}
   \FOR {$d{=}1$ to $d_{max}$}
      \FOR {each $Sour_j$}
         \FOR {each $sour_{j,k} \in Sour_j$}
            \STATE {$X_{i,j,k} \gets$ simplex, using $vert_i, pls_i, sour_{j,k}$}
            \FOR {each vertex pair in $X_{i,j,k}$}
               \STATE {compute edge distance, using $Met$}
            \ENDFOR   
         \STATE {$cont_{i,j,k} \gets$ content of $X_{i,j,k}$, using $CM$}
         \ENDFOR   
      \ENDFOR
      \STATE {$e_d \gets$ $argextreme_{j,k}$ of $cont_{i,j,k}$, using $Pol$}
      \STATE {$names_d \gets$ names of $sour_{j,k}$, using $\langle e_d \rangle$}
      \STATE {$pls_i \gets pls_i$ concat $names_d$}
   \ENDFOR
\ENDFOR

\end{algorithmic}
\end{algorithm}

\subsection{Empirical Properties of Pseudo-labels}
\label{subsub:empirical}

In what follows, we have used VGG16 trained on Imagenet1K
as classifier $f$, and we have partitioned the $Full$ ImageNet22K dataset into
a $Part$ collection of the 16 non-intersecting semantic subsets given
in Sec.~\ref{sub:motivation}.  
We will now
refer to policies without angle brackets or commas; 
$\langle C, f, f, f \rangle$ becomes simply $\T{Cfff}$.

\textbf{Outliers and tractability.}
We note that a few of our 16 mean vectors, particularly $fungus$,
$sport$, and $furniture$, repeatedly show up as outlier vertices in
the polytopes under construction, as suggested by their relations shown in Fig.~\ref{fig:confusion16}.  They therefore tend to occur early
in the output pseudo-label sequences.


We observe empirically, as suggested in~\cite{d2005cayley}, that the
search for these maximal and minimal simplicies has to be done exhaustively.
For example, through exhaustive search on
our test dataset, we find that the $1$-simplex with minimal content is
$\langle plant, tree\rangle$, yet the minimal $2$-simplex is $\langle
fabric, garment, person \rangle$, and then the minimal $3$-simplex is
$\langle fabric$, $garment$, $plant$, $tree \rangle$.

\begin{figure*}[!ht] 
\begin{subfigure}{.41\linewidth}
  \centering
  \includegraphics[width=\linewidth]{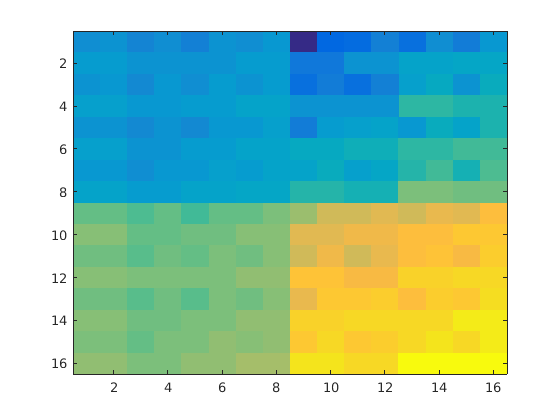}
  \caption{Entropy of all 256 policies for $d{=}4$,
  from $\T{cccc}$ to $\T{FFFF}$.}
\end{subfigure}
\hspace{0.07\linewidth}
\begin{subfigure}{.45\linewidth}
  \centering
  \begin{equation*} 
  \begin{smallmatrix}
   &            &            &            &            &    &            &            &            &            \\
 1 & \T{cccc  } & \T{cccf  } & \T{\cdots} & \T{cfff  } & \, & \T{fccc  } & \T{fccf  } & \T{\cdots} & \T{ffff  } \\
 2 & \T{cccC  } & \T{cccF  } & \T{\cdots} & \T{cffF  } & \, & \T{fccC  } & \T{fccF  } & \T{\cdots} & \T{fffF  } \\
 3 & \T{ccCc  } & \T{ccCf  } & \T{\cdots} & \T{cfFf  } & \, & \T{fcCc  } & \T{fccF  } & \T{\cdots} & \T{ffFf  } \\
   & \T{\vdots} & \T{\vdots} & \T{\ddots} & \T{\vdots} & \, & \T{\vdots} & \T{\vdots} & \T{\ddots} & \T{\vdots} \\
 8 & \T{cCCC  } & \T{cCCF  } & \T{\cdots} & \T{cFFF  } & \, & \T{fCCC  } & \T{fCCF  } & \T{\cdots} & \T{fFFF  } \\[2ex]
 9 & \T{Cccc  } & \T{Cccf  } & \T{\cdots} & \T{Cfff  } & \, & \T{Fccc  } & \T{Fccf  } & \T{\cdots} & \T{Ffff  } \\
10 & \T{CccC  } & \T{CccF  } & \T{\cdots} & \T{CffF  } & \, & \T{FccC  } & \T{FccF  } & \T{\cdots} & \T{FffF  } \\
11 & \T{CcCc  } & \T{CcCf  } & \T{\cdots} & \T{CfFf  } & \, & \T{FcCc  } & \T{FcCf  } & \T{\cdots} & \T{FfFf  } \\
   & \T{\vdots} & \T{\vdots} & \T{\ddots} & \T{\vdots} & \, & \T{\vdots} & \T{\vdots} & \T{\ddots} & \T{\vdots} \\
16 & \T{CCCC  } & \T{CCCF  } & \T{\cdots} & \T{CFFF  } & \, & \T{FCCC  } & \T{FCCF  } & \T{\cdots} & \T{FFFF  } \\
   &  1         &  2         &            &  8         &    &  9         &  10        &            & 16         \\
  \end{smallmatrix}
  \end{equation*}
  \caption{Policies corresponding to the heatmap at left.}
\end{subfigure}
\caption{Heatmap showing the entropy of the pseudo-labels generated for $d{=}4$ policies applied to downsampled ImageNet22K data. The diagram is fractal.}
\label{fig:entropy256}
\end{figure*}

\textbf{Impact of the first policy.}
We illustrate an important statistical property of the
resulting pseudo-labelings
in the heatmap of Fig.~\ref{fig:entropy256}, which displays
the entropy of pseudo-labels generated for $Part$ under the 256
possible policies of order $d{=}4$.

What is visually apparent is that the variability depends primarily on
the initial policy in the sequence.  The left half of the diagram
shows policies that begin with $\T{c}$ or $\T{C}$ (first policy
decision is ``closest''); the right half begins with $\T{f}$.or
$\T{F}$ (first policy decision is ``farthest'').  The top half shows
policies that begin with $\T{c}$ or $\T{f}$, which produce a single
source name at $d{=}1$; the bottom half begins with $\T{C}$ or
$\T{F}$, which produce two source names at $d{=}1$.

The diagram shows that diversity of pseudo-labels increases roughly
from upper left, $\T{cccc}$, to lower right, $\T{FFFF}$.  The general
progression is $\T{c}, \T{f}, \T{C}, \T{F}$, in the pattern of
$\begin{bmatrix} \T{c} & \T{f} \\ \T{C} & \T{F} \end{bmatrix}$.
Closer examination shows that the diagram is fractal, and that it follows this pattern at all scales.

The major anomaly is $\T{fccc}$, at position $(1,9)$, colored strongly
dark, which can be interpreted as ``apply the $4$-simplex consisting
of the worst outlier source and its three nearest neighbors''---which
for many different inputs is exactly the same, giving minimal entropy.
In contrast, the rightmost column of the diagram, which traces
policies from $\T{ffff}$ at $(1,16)$ to $\T{FFFF}$ at $(16,16)$, shows a
monotonic and nearly linear increase in entropy to the global maximum.

\textbf{Growth and predictability.}
The G2L algorithm creates pseudo-label sequences that are combinations of
source names.  The number of potential sequences of any given length
can be very large, for example, $\binom{16}{4}{=}1,820$,
$\binom{16}{6}{=}8,008$ and $\binom{16}{8}{=}12,870$.  However, when
the algorithm was applied to the ImageNet22K training dataset,
the number of unique sequences were typically much less, as
shown in Fig.~\ref{tbl:pseudo-dataset}.  
Even $\T{FFFF}$ of length 8 in the prolific extreme lower right
corner produced only 2,643.  This slower growth reflects the
non-random correlated semantic clustering of the data.  

\section{Experimental Evaluation}
\label{sec:experiment}

\begin{figure*}[!ht]
\begin{subfigure}{.38\linewidth}
    \centering
    \begin{tabular}{||l| r||l| r||}
    \hline
    \hline
    Policy & \#P-L & Policy & \#P-L\\ 
    \hline
    \hline
    \texttt{cccc}*   &  40 & \texttt{Cfff}*    & 526  \\
    \hline
    \texttt{cccccc}* & 103 & \texttt{Fccc}     & 543  \\
    \hline
    \texttt{fccc}    & 127 & \texttt{vanilla}* & 1000 \\ 
    \hline
    \texttt{cfffff}* & 159 & \texttt{Ffff}*    & 1121 \\
    \hline
    \texttt{C}       & 201 & \texttt{FCff}     & 1390 \\
    \hline
    \texttt{ffff}    & 225 & \texttt{FCCC}*    & 1807 \\
    \hline
    \texttt{Cccc}    & 345 & \texttt{FFFF}     & 2643 \\
    \hline
    \texttt{CCff}    & 502 &                  &      \\
    \hline
    \hline
    \end{tabular}
    \caption{The 15 different base model datasets; ``\#P-L'' is ``Count of Pseudo-Labels''.  The 7 models with asterisks were those used for transfer learning experiments.}
    \label{tbl:pseudo-dataset}
\end{subfigure}
\hspace{0.04\linewidth}
\begin{subfigure}{.58\linewidth}
  \centering
  \includegraphics[width=\linewidth]{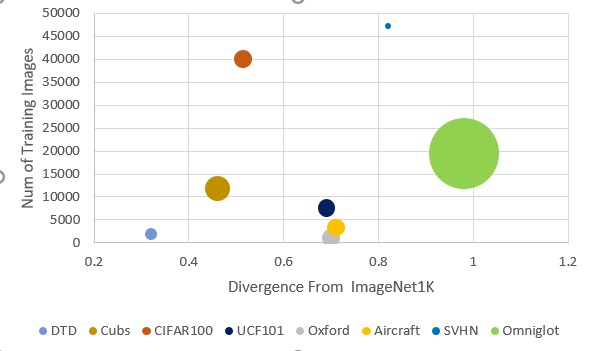}
  \caption{Bubble chart showing the variety of the 8 target datasets evaluated. Size of bubble indicates number of labels. X is divergence from ImageNet1K, and Y is the number of training images}
  \label{fig:targetdiv}
\end{subfigure}
\caption{(a) Base models and (b) Target datasets}
\label{fig:table1_figure4}
\end{figure*}

Using our geometric technique, 
we created a number of pseudo-labeled datasets for the images in ImageNet1K, as shown in Fig.~\ref{tbl:pseudo-dataset}. We then trained ResNet27 models using six pseudo-labeled datasets, creating base models for further transfer learning. These six were: $\texttt{cccc}$, $\texttt{Cfff}$, $\texttt{Ffff}$, $\texttt{FCCC}$, $\texttt{cccccc}$, $\texttt{cfffff}$. These six were chosen because they represent a broad spectrum of unique label counts, they explore policies starting with different initial extremizing decisions, and they show the effect of increased dimensions. 
We observe that the CFA policy introduced in~\cite{Dube_2019} can be obtained from our approach; by our notational convention it would be, simply, policy $\texttt{C}$.

We also created a baseline model using the vanilla ImageNet1K dataset of images and human-annotated labels. This model attained a top-1 average accuracy of 66.6\%, which is suitable for a ResNet27 model.  The same hyperparameters and training setup was used for all the pseudo-labeling models. 
We chose ResNet27 because residual networks~\cite{resnet27} are considered state of the art,
and ResNet27 is easy to train, while still being large enough for
our datasets.

To evaluate the usefulness of these base models, we focused on eight target workloads taken from the Visual Domain Decathlon~\cite{odeca} and other fine-grained visual classification tasks. 
The choice of target datasets was made to have sufficient diversity in terms of number of labels, number of images, number of images per label, and divergence with respect to ImageNet1K.  Divergence is here computed by first normalizing the representative vectors of each dataset so that their components (which are all non-negative) sum to 1, then applying the usual Kullback–Leibler divergence formula~\cite{kullback1951}.

Since we want to compare the performance of pseudo-labeling with respect to vanilla ImageNet1K, we selected only those datasets whose transfer learning accuracy under vanilla were not close to 1. This ensures that the comparison with vanilla is not trivial (otherwise, all policies also have accuracies very close to 1). The target workloads evaluated included Aircraft~\cite{oaircraft}, CIFAR100~\cite{ocifar100},  Describable Textures (DTD)~\cite{odtd}, Omniglot~\cite{oomni}, Street View House Number (SVHN)~\cite{osvhn}, UCF101~\cite{oucf}, Oxford VGG Flowers~\cite{ovgg}, and Caltech-UCSD Birds (CUBS)~\cite{ocubs}. These span a range of divergence from ImageNet1K, and possess different labels and dataset sizes, as shown in Fig.~\ref{fig:table1_figure4}.

These target workloads were then learned from pseudo-labeled and human-annotated (vanilla ImageNet1K) source models over five different learning rates. The inner layers were set to learning rates ranging over 0.001, 0.005, 0.010, 0.015, and 0.020, and the last layer was set to a learning rate ten times that. 

 Each source model was trained using Caffe1 and SGD for 900K iterations, with a step size of 300K iterations, an initial learning rate of 0.01, and weight decay of 0.1. The target models were trained with identical network architecture but with a training method with one-tenth of iterations (90K) and step size (30K). A fixed random seed was used throughout all training.
Thus a total of 280 transfer learning experiments (8 targets $\times$ 7 policies $\times$ 5 learning rates), with same set of hyperparameters, were conducted and compared for top-1 accuracy. 

\section{Observations}
\label{sec:observations}

\begin{figure*}[!ht]
\begin{subfigure}{.38\linewidth}
    \centering
    \begin{tabular}{||c|c|c|c||}
    \hline
    \hline
    Dataset & Div. & Vanilla & G2L\\          
    \hline
    \hline
    DTD & 0.32 & \bf{0.4388} &            0.4282\\
    \hline
    CUBS & 0.46 & \bf{0.3485} &   0.2825\\
    \hline
    CIFAR100 & 0.51 & \bf{0.7320}	&          0.7065\\
    \hline 
    UCF101	& 0.69 & 0.7500 &   \bf{0.7546}\\	
    \hline
    Oxford & 0.70 & 		0.7585& 		  \bf{0.7628}\\
    \hline
    Aircraft & 0.71 & \bf{0.4882}&               0.4744\\
    \hline
    SVHN & 0.82 &	0.9351 & 	          \bf{0.9385}\\
    \hline
    Omniglot & 0.98  & 0.7961	&          \bf{0.7975} \\
    \hline
    \hline
    \end{tabular}
    \caption{The 8 targets tested, their divergence from ImageNet1K, and their accuracies.
    Accuracies are the maximum over 5 learning rates and 6 policies.}
    \label{tbl:overall_accuracy}
\end{subfigure}
\hspace{0.02\linewidth}
\begin{subfigure}{.58\linewidth}
  \centering
  \includegraphics[width=\linewidth]{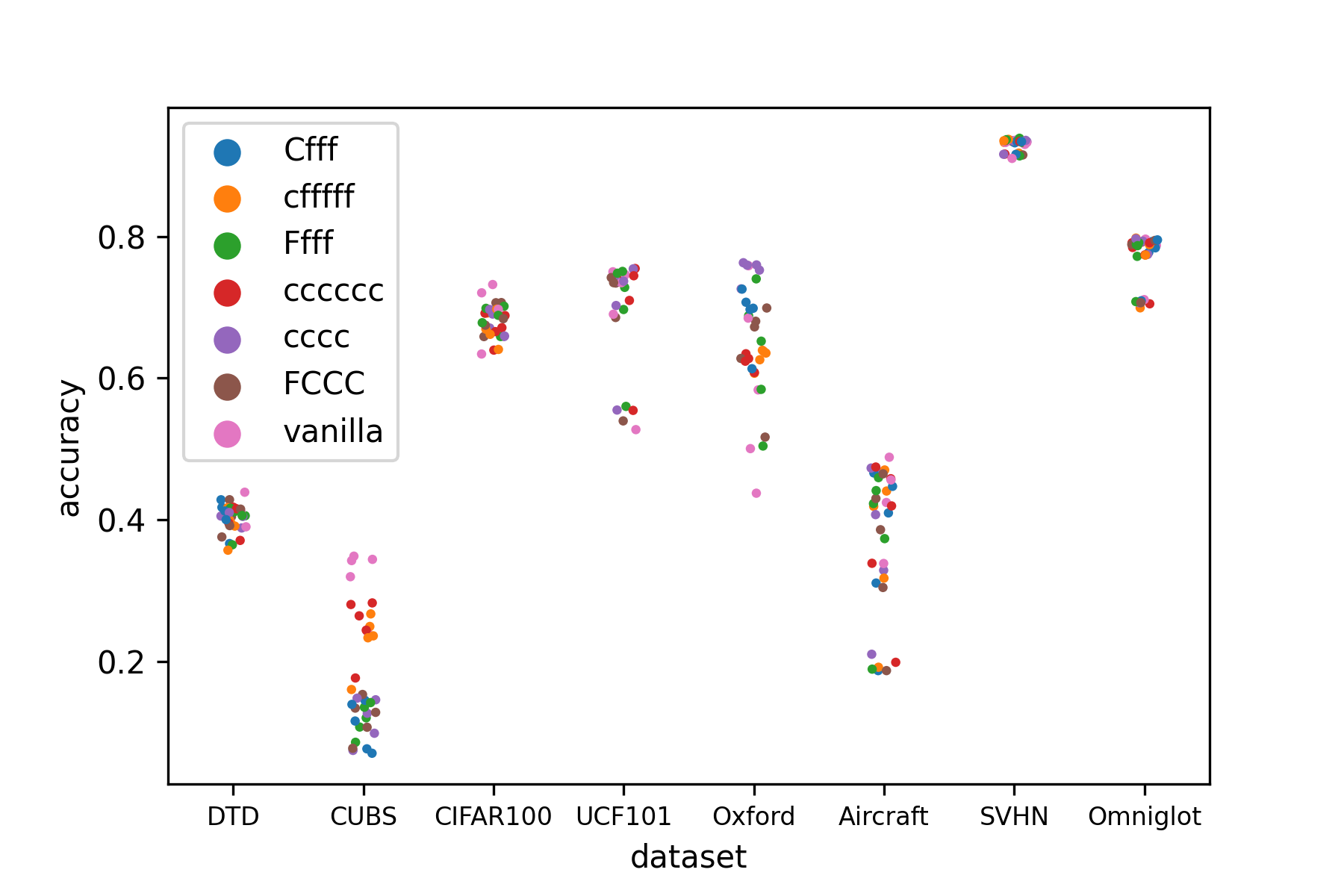}
  \caption{Performance of 6 G2L policies compared to vanilla, on 8 targets at 5 different learning rates.}
  \label{fig:winloss1}
\end{subfigure}
\caption{(a) Divergence versus accuracy for 8 targets. (b) All 280 experiments.}
\label{fig:figure5_table2}
\end{figure*}

\textbf{Overall Accuracy.}
Figs.~\ref{tbl:overall_accuracy} and \ref{fig:winloss1}
 compare the transfer learning top-1 accuracy of vanilla with our pseudo-labeling approach.  
Fig.~\ref{tbl:overall_accuracy} shows that the divergence measure closely tracks the accuracy of both vanilla and G2L, with a correlation coefficient of 0.7.  For datasets whose divergence is above 0.6, G2L beats vanilla four times out of five.  
Our approach 
outperforms vanilla in four high divergent cases (Omniglot, SVHN, Oxford, and UCF101). 
For the other four cases (DTD, Cubs, CIFAR100, and Aircraft) where it underperforms, its performance was very similar to vanilla. 
Taken together, the average of all eight winners, 
compared to the average of just vanilla, decreases the overall error rate by 0.43\%.  

\begin{figure}[!ht]
\begin{subfigure}{.3\linewidth}
  \centering
  \includegraphics[height=1in,width=\linewidth]{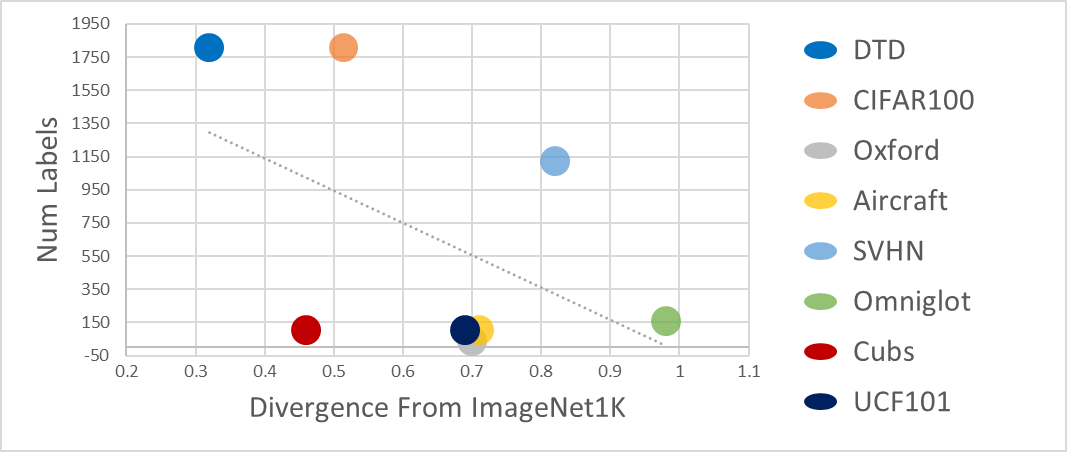}
  \caption{}
  \label{fig:winloss2}
\end{subfigure}
\hspace{0.02\linewidth}
\begin{subfigure}{.3\linewidth}
  \centering
  \includegraphics[height=1in,width=\linewidth]{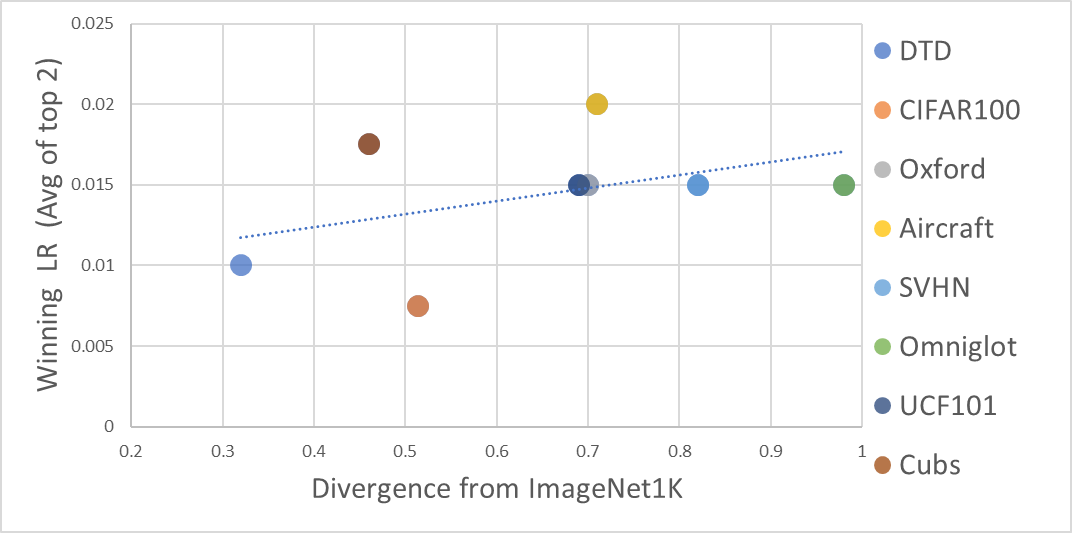}
  \caption{}
  \label{fig:winloss3}
\end{subfigure}
\hspace{0.02\linewidth}
\begin{subfigure}{.3\linewidth}
  \centering
  \includegraphics[height=1in,width=\linewidth]{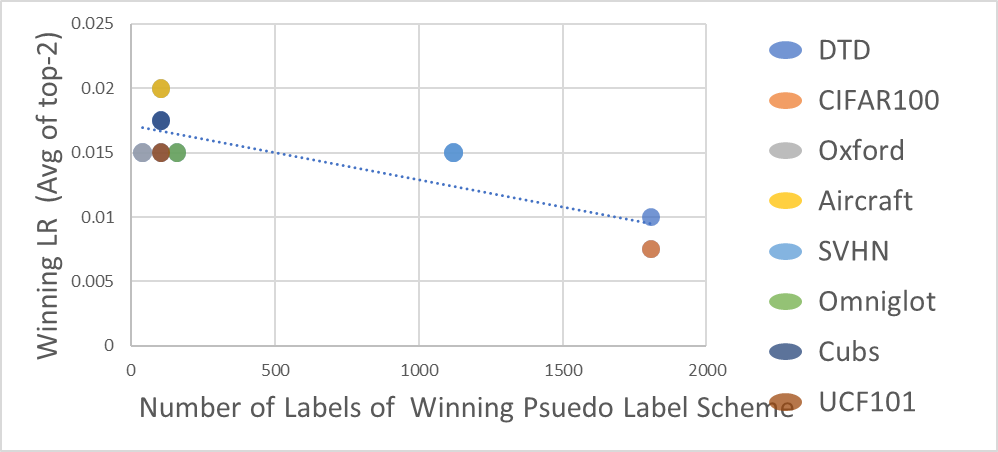}
  \caption{}
  \label{fig:winloss4}
\end{subfigure}
\caption{(a) Divergence vs. Number of Labels. (b) Divergence vs. Learning Rate. (c) Number of Labels vs. Learning Rate.}
\label{fig:figure6_7_8}
\end{figure}

\textbf{Divergence vs. Number of Labels.} 
As divergence from Imagenet1K increased for the targets, pseudo-labeling schemes with a lesser number of unique labels performed better; see Fig.~\ref{fig:figure6_7_8}(a). In cases where pseudo-labeling schemes did better than the vanilla ImagNet1K labels, the label count was 40 to 160, in contrast to 1,000 in the vanilla. 

\textbf{Divergence vs. Learning Rates.}
As divergence from Imagenet1k increased for the targets, higher learning rates were better for the pseudo-labeling schemes; see Fig.~\ref{fig:figure6_7_8}(b). In the cases where pseudo-labeling schemes did better than vanilla ImageNet1K, learning rates of 0.015 did best. Even for the vanilla ImageNet1K labels, a higher learning rate was generally better for high divergent workloads.

\textbf{Number of Labels vs. Learning Rates.}
Pseudo-labeling schemes with fewer unique labels performed better at higher learning rates; see Fig.~\ref{fig:figure6_7_8}(c). In cases where pseudo-labeling schemes did better than vanilla, high learning rates like 0.015 were best.

\section{Limitations and Future Work}
\label{sec:limitations}

\textbf{Theory Limitations.}
The initial partitioning of the ImageNet22K dataset into 16 categories is currently heuristic. 
It known that the use of the second-last vector as a representative is not helpful if incoming data is significantly different on the signal level~\cite{DBLP:journals/corr/Garcia-GasullaP17}, and that vectors of other layers are.
The aggregation method to aggregate the representative vectors of the sources relies on a simple mean.
An efficient method for selecting the probable best policy is currently unexplored.

\textbf{Theory Future Work.}
Initial partitioning could be cast as an (approximate) optimization problem, as can the selection of representative processing levels.  
Simple aggregation by a mean vector can be augmented by a multi-vector clustering technique.  
Inter-cluster distance measures like ``energy distance''~\cite{baringhaus2004new} would then be used instead of Euclidean.
More powerful intermediate $n^{th}$ order Voronoi methods should be explored; see Fig.~\ref{fig:voronoi}(b).  
The space of policies over many problems should be examined in order to determine heuristics about policy behaviors, particularly those regarding the likelihood of accuracy improvement.
The G2L methods needs to be applied to data modalities other than vision. 

\textbf{Practice Limitations.}
How well a given initial partition spans the available high-dimensional feature space has not been quantified.
Learning rates for each layer in these experiments were identical.  
Policies were applied according to a fixed script.
The value of $d_{max}$ is a completely free meta-parameter.

\textbf{Practice Future Work.}
A method for optimizing the initial partition according to some figure of merit would be useful, particularly since human-annotated labels are quite sensitive to fine details such as color, texture, shape, etc.   
A more appropriate selection of learning rates for different layers of network can significantly improve accuracy during fine-tuning~\cite{DBLP:journals/corr/abs-1807-11459}; 
a thorough exploration, coordinated with knowledge of specific policy strengths and weaknesses, should be attempted.
Search techniques other than greedy should be explored, that more intelligently select the best policy to execute next, and that stop without requiring a value for $d_{max}$.
These G2L approach, particularly since it generates descriptive pseudo-labels, can and should be applied to the problem of data augmentation, and to human label error correction.

\section{Conclusion}
\label{conclusion}

We have demonstrated a novel but general technique for automatically creating descriptive content-aware pseudo-labels for transfer learning,
based on a classical result from the geometry of high dimensions. 
It computes hypervolume content as a heuristic for label quality.
The expected accuracy is approximately forecast by the divergence between the representative vectors of sources and targets.  
Our 280 experiments show that this mechanical technique generates base models that have similar or better transferability, compared to usual methods.
This G2L approach overall increases classification accuracy, particularly on incoming datasets that are unusually divergent from the human-labeled training set.
Our novel geometry-based pseudo-labeling method can be extended in several theoretic and practical ways, and can be applied to other data modalities.

\clearpage
\bibliographystyle{splncs04}
\bibliography{references}

\end{document}